# Skeleton based Activity Recognition by Fusing Part-wise Spatio-temporal and Attention Driven Residues

Chhavi Dhiman[1], Dinesh Kumar Vishwakarma[2], Paras Aggarwal[3]
Biometric Research Laboratory, Department of Information Technology,
Delhi Technological University, Delhi-110042
Email: chhavi1990delhi@gmail.com[1], dinesh@dtu.ac.in[2] , paraspro2020@gmail.com

**Abstract:**
There exist a wide range of intra-class variations of the same actions and inter-class similarity among the actions, at the same time, which makes the action recognition in videos very challenging. In this paper, we present a novel skeleton-based part-wise Spatio-temporal CNN – RIAC Network-based 3D human action recognition framework to visualise the action dynamics in part wise manner and utilise each part for action recognition by applying weighted late fusion mechanism. Part-wise skeleton-based motion dynamics helps to highlight local features of the skeleton which is performed by partitioning the complete skeleton in five parts- Head to Spine (HS), Left Leg (LL), Right Leg (RL), Left Hand (LH), Right Hand (RH). The RIAFNet architecture is greatly inspired by the InceptionV4 architecture which unified the ResNet and Inception based Spatio-temporal feature representation concept and achieving the highest top-1 accuracy till date. To extract and learn salient features for action recognition, attention driven residues are used which enhance the performance of residual components for effective 3D skeleton-based Spatio-temporal action representation. The robustness of the proposed framework is evaluated by performing extensive experiments on three challenging datasets such as UT Kinect Action 3D, Florence 3D action Dataset, and MSR Daily Action3D datasets, which consistently demonstrate the superiority of our method.

## 1. Introduction

Human action recognition in videos is an essential field of computer vision and has an extensive range of applications i.e. Human-robot vision, human-computer interaction, intelligent video surveillance, game control, elderly health care, sports etc. [1] [2] [3]. Traditional researches in this field initially referred to RGB video-based action recognition. The information recorded in these videos is one projection of the entire scene, along with the motion content of the object in action, on a particular plane. However, the motion characteristics of the human body are well represented in the 3D space. Considering this fact human bodies are, now, modelled as a set of 20 or 25 skeleton joints in 3D coordinate system and rigid bones. Position variation of the skeleton joints in 3D space are responsible for human actions. Currently, the world skeleton joint coordinates can be extracted using the cost-effective depth sensor combining the real-time skeleton estimation algorithms [4], [5]. It has resulted in significant amount of work, coming up by exploring effective approaches for skeleton-based action recognition. Human skeleton-based action recognition approaches generally exploit temporal dynamics of the action [6], by developing explicit temporal dynamics model such as Fourier Temporal Pyramids (FTPs) [7] [8] and probabilistic graphical model- HMMs [9] . FTP based methods utilize a limited amount of contextual information constrained by the size of the time windows. Hence, failed to capture the temporal sequences of actions, globally. Alignment of temporal sequences in HMM and obtaining a corresponding emission distributions is tough. Recently, several Recurrent Neural Networks (RNNs) models [10] and Long-Short Term Memory (LSTMs) models [11] are proposed to learn the temporal dynamics of action features. A large amount of information lies in the spatial distribution of human poses [12] [13]. Well-designed spatial-temporal encodings of skeleton sequences are more effective than only temporal information of skeleton-based action representation. Therefore, the methods [14] [15] [16] are proposed which process both spatial and temporal information of the action together, leading to more effective recognition systems. Pham et al. [14] [15] transformed the body joint coordinates into 3D arrays which can capture the spatial-temporal evolutions of 3D motions from skeleton sequences and processed as images using Deep Residual Neural Networks (ResNets) to learn and recognize human action from skeleton data provided by Kinect sensor.

In this work, we aim to take full advantages of spatio-temporal CNN to build an end-to-end learning framework for HAR by utilising both spatial and temporal information of action using 3D skeleton based representations of the actions. The concept of InceptionV4 [17] architecture has evolved from InceptionV1/ GoogleNet [18] and later fused with the concept of ResNet [19]. It greatly enhanced the recognition performance of the InceptionV4 [17] architecture from other versions of Inception network with the minimum top-1 error. However, in order to achieve superior performance than other state-of-the-art, use of InceptionV4 [17] network significantly demands a more significant number of parameters and operations. It greatly inspired us, to utilise the concept of inception blocks and residues, in designing the proposed RIAC-Net architecture, for feature extraction. The proposed work defines a novel action descriptor by combining attention based residues with Spatial-Temporal based Convolution Features (STCF). The main contribution of the proposed work are four folds: -

- A novel part wise Spatio-temporal CNN based action descriptor –RIAC-Net is designed, using attention driven residual concept. The complete skeleton is partitioned into five significant parts: a) head to spinal (HS), b) left leg (LL), c) right leg (RL), d) left hand (LH), and e) right hand (RH).
- To learn the temporal dynamics of the complete action, the defined part-wise action descriptors, are learnt using two consecutive layers of Long Short Term Memory (LSTM) units. And the final prediction for the test action is taken by using a weighted fusion of five predictions computed from each part of the skeleton.
- To exhibit the effect of the engineering part-wise skeleton-based RIAC-Net action description over complete skeleton-based action description, we compared the obtained results of the proposed action recognition framework with the complete skeleton based RIAC-Net action descriptor results. And it is observed that part-wise RIAC-Net action descriptor performs superior.
- Finally, the effectiveness of the proposed framework for skeleton-based action recognition is shown, by achieving the state-of-the-art performance on three benchmark datasets such as UT Kinect Action 3D, Florence 3D actions Dataset and MSR Daily Action3D datasets.

Rest of the paper is organised as follows: - Section 2 discusses related work. In section 3 the details of the proposed work are illustrated. Datasets and the experiments are described in Section 4. Section 5 describes the experimental results. In the last, a conclusion is drawn about the proposed work and future work is discussed, in section 6.

**2. Related Work**

The skeleton-based action recognition approaches [20] [21] are progressing towards the temporal dynamics of the action using RNNs and LSTMs, recently. Du et al. [20] proposed an end-to-end hierarchical RNN to encode the relative motion between skeleton joints. Skeletons were split into anatomically relevant parts, which were fed into each independent subnet to extract local features. Shahroudy et al. [21] introduced a part-aware LSTM which possess part-based memory sub-cells and a new gating mechanism, showing the superior performance of LSTM over some hand-crafted features and RNN. To learn the human motion features of the skeleton sequence, RNN-LSTM [22] allows the network to access and store the long-range contextual information of a skeleton sequence. Several authors [23] [24] exploited feature learning ability of CNNs which largely focused on a better skeletal representation and learning with simple CNNs. To better capture the Spatio-temporal dynamics of the skeleton sequences, some authors [25] [26] [27] used CNN as a spatial feature extractor and unified with RNN-LSTM network to model human motion. However, it is noticed that RNN-LSTM based approaches performed better. On the other side, the use of RNNs results in overfitting if number of input features are short enough to train the network and computational time dynamically increases with the number of input vectors.

Well-designed spatial-temporal encodings of skeleton sequences are more effective than only temporal information of skeleton-based action representation. Tu et al. [28] defined the correlation among three-dimensional signal using 3DCNN to capture spatial and temporal information of the action sequence. Liu et al. [29] mapped the skeleton joints in 3D coordinate space before extracting view-invariant Spatio-temporal features, which significantly improved the action recognition results. Whereas, the work [30] learnt adequate geometric features of 3D human actions by using Lie Group and unified it with deep neural networks. Chen et al. [31] encoded the skeleton joints as part based 5D feature vector to identify the most relevant joints of skeleton during the action sequence using a two-level hierarchical framework. Amor et al. [32] used trajectories on Kendall's shape manifolds to model the dynamics of human skeleton poses and used a parametrization-invariant metric for aligning, comparing, and modelling skeleton joint trajectories, to deal with noise caused by different execution rates of the actions performed by humans. However, this method is very time-consuming.

A good amount of work is also done to address the spatial representation of human skeleton poses which is mainly driven by the fact that an action can be characterized properly by the interactions or combinations of a subset of skeleton joints [22]. The methods to model action spatial patterns can be categorised in two classes: part-based model and sub-pose model. In the first category of spatial pattern modelling, the skeleton is divided into several groups, instead of considering the complete skeleton. The HBRNN [33] model decomposed the skeletons into five parts, two arms, two legs, and one torso, and built a hierarchical recurrent neural network to model the relationship among these parts. Similarly, Shahroudy et al. [21] proposed a part-aware LSTM model that construct the relationship between body parts. Whereas, in sub-pose model, the informative joints or their interactions are mainly focused. A handcraft features based approach [34] defined a SMIJ model which select the most informative joints by calculating statistical parameters such as mean and variance of joint angle trajectories and used the sequence of selected informative joints to represent the action. Wang et al. [35] mined co-occurrence distinctive spatial body-part structures as spatial part-sets and temporal evolutions of pose as temporal part sets. Whereas Lillo et al. [36] learnt the spatio-temporal annotations of complex actions using motion poselets and actionlet dictionaries. Such annotations help to understand that which body part is active for a particular action but not discriminative enough in classification.

## 3. Proposed Work

In this section, the proposed framework for skeleton-based human action recognition is explained. It describes the formation of Compact Action Skeleton Sequence formation and proposed RIAC-Net architecture design in detail which includes three main steps- formation of Spatial-Temporal Convolution Features (STCF), defining Attention Driven Residual Block (ADRB), and lastly learning part-wise RIAC-Net based action features and to ensemble the predictions per part using weighted fusion scheme.

### 3.1 Compact Action Skeleton Sequence (CASS) generation

CASS is basically, a projection of each frame 3D coordinates of skeleton joints on the image frame which describes the spatial variation of the human skeleton pose during the action. The temporal details about the sequence of human poses are encrypted by using different colour coding for skeletons in such a way that colour of the skeletons change with time to exhibit the sequence of occurrence of frames. To exploit the discriminative local features of the actions, the generated CASS are further divided into five significant parts: $i$) Head to Spine ($HS$) $ii$) Left Leg ($LL$) $iii$) Right Leg ($RL$) $iv$) Left Hand ($LH$) and $v$) Right Hand ($RH$). Therefore, CASS is defined for FS as well as for other five parts of the skeleton - {HS, RL, LL, RH, LH}. Let an action video has $n$ no. of frames $\{f_1, f_2 \ldots \ldots, f_n\}$, and each frame possess a human skeleton with $k$ no. of skeleton joints $i.e.$ $\{(J_{x1}, J_{y1}, J_{z1}), (J_{x2}, J_{y2}, J_{z2}), \ldots \ldots (J_{xk}, J_{yk}, J_{zk})\}$, $k\epsilon[1,20]$. According to the configuration of skeleton joints in Figure 1, skeleton joints are partitioned in five groups as: $HS = \{J_4, J_3, J_2, J_1\}$, $LL = \{J_{13}, J_{14}, J_{15}, J_{16}\}$, $RL = \{J_{17}, J_{18}, J_{19}, J_{20}\}$, $LH = \{J_5, J_6, J_7, J_8\}$, $RH = \{J_9, J_{10}, J_{11}, J_{12}\}$. To generate CASS, the variations in joint coordinates of each group are sketched, mathematically defined as Eq. (1):

$$CASS_p = \begin{pmatrix} J_k^t & \cdots & J_k^n \\ \vdots & \ddots & \vdots \\ J_{k+4}^t & \cdots & J_{k+4}^n \end{pmatrix} \quad (1)$$

where $k$ is the first skeleton coordinate number of each group, $p$ is the partition label $\exists\ p\epsilon(HS, LL, RL, LH, RH)$ n is the number of frames in the action video.

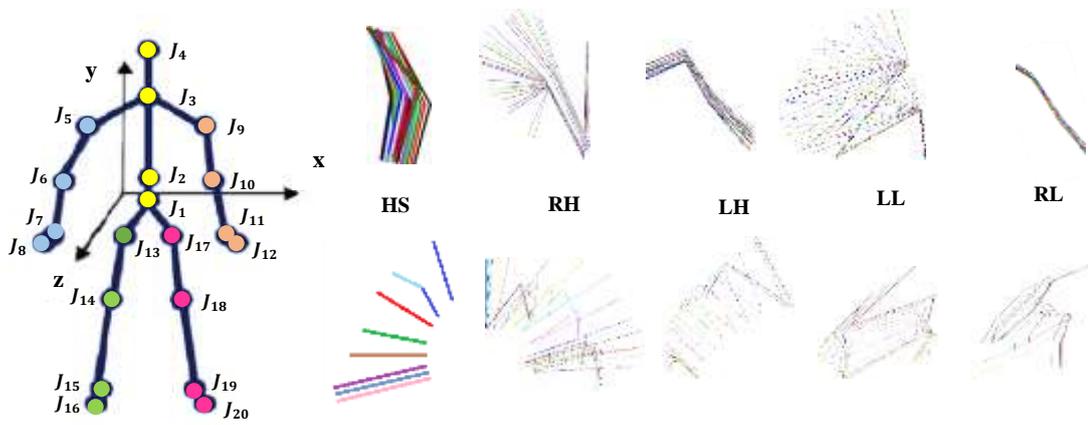

Fig. 1(a) Configuration of skeleton joints. Head to Spine (HS) joints is colored in Yellow, LH joints are colored in Blue, RH joints are coloured in Brown, LL joints are coloured in Green, RL joints are coloured in Pink (b) CASS formation for two actions – 'waving hand' and 'sitting down', of the Florence3D Action dataset

From the sample images of part-wise CASSs, for two actions- waving hand and sitting down, of the Florence3D Action dataset, in Fig. 1, it is observed that different amount of motion is associated with each part of the skeleton for an action resulting in unique patterns for each action. The part wise feature extraction and learning highlight the local dynamics of the skeleton. However, if the complete skeleton is processed to extract spatial deep features, the prominent movements in the action would be subsided.

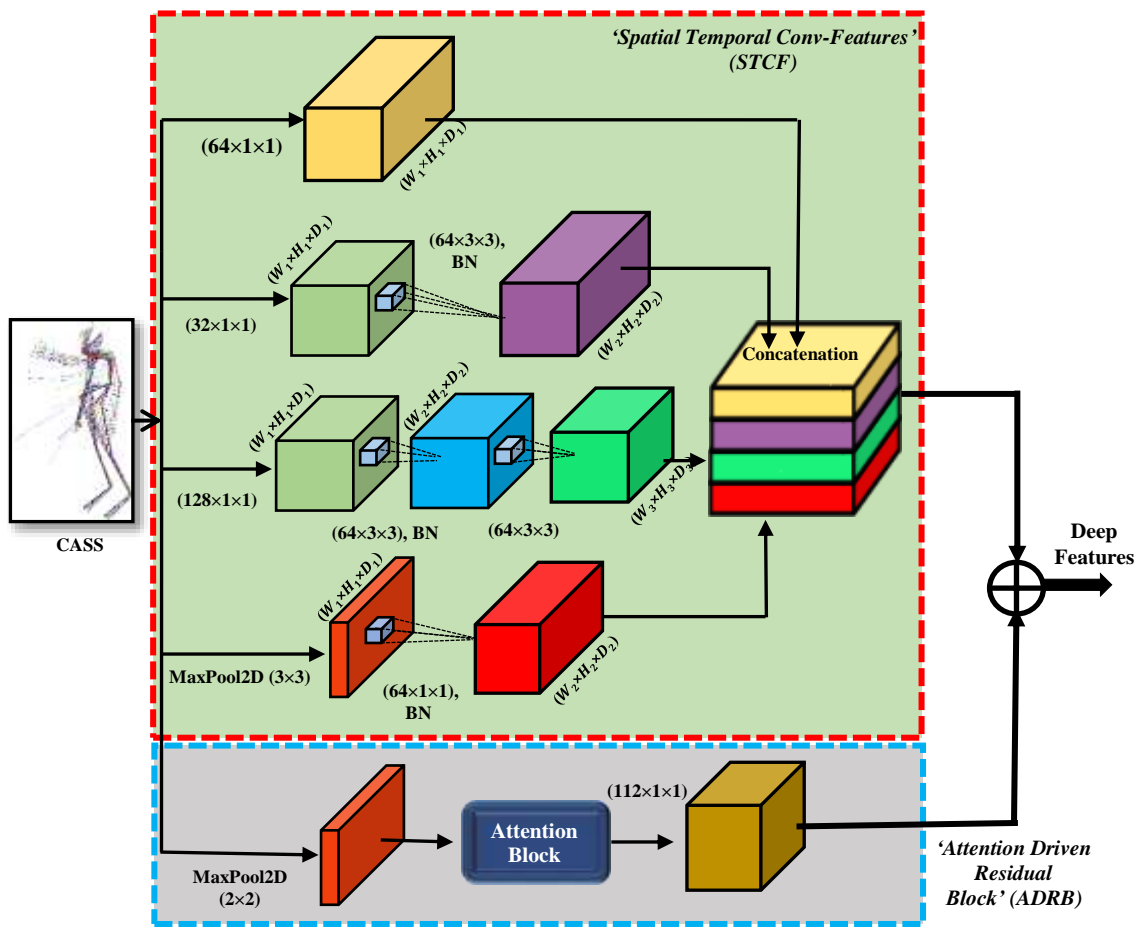

Fig. 2. Proposed Residual Inception Attention-based Convolution Network (RIAC-Net) block diagram

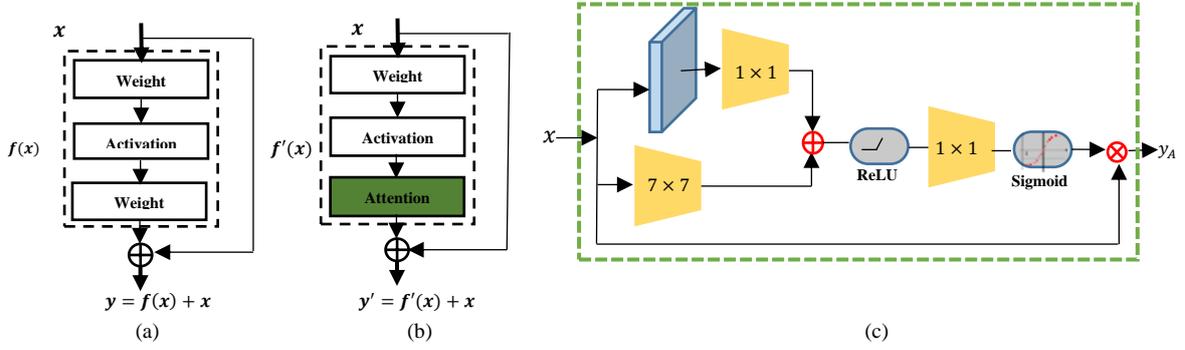

Fig. 3. Illustration of Attention Driven Residual Block architecture (a) Basic Residual Block (b) Attention Driven Residual Block (c) Attention block

### 3.2 Skeleton based action recognition with RIAC-Net

The training of Inception networks with residual connections has accelerated significantly, resulting in outperforming the similarly expensive inception networks without residual connections [17]. Therefore to solve the problem of skeleton-based action recognition for large inter-class similarity Residual Inception Attention-based Convolution Network (RIAC-Net), Fig. 2, is designed which is majorly divided into two parts- 'Spatial-Temporal Convolution Features' (STCF) and 'Attention Driven Residual Block' (ADRB).

#### A. Spatial-Temporal Convolution Features (STCF)

Salient features in an image, generally, can have an extremely large variation in size i.e. covering major section of the image or small section. Convolution helps to recover Spatio-temporal features only with right selection of kernel size. A large convolution kernel has large receptive field that highlights the globally distributed information and a smaller convolution kernel is preferred for locally distributed information. Use of multiple sized kernels in convolution filters i.e. $(1 \times 1, 3 \times 3, 5 \times 5)$ is an efficient solution to select the appropriate kernel size for good convolution features [18]. It essentially widens the network size and also computationally less expensive than deeper networks. It is the basic concept behind building the inception blocks [18] that targeted large size variations of spatial features. The convolution filters are made computationally more efficient by factorising $(5 \times 5)$ filters with two $(3 \times 3)$ filters in STCF. Description of the parameters of STCF block is provided in Table 1. It can be notified that equal-sized features $(W(112) \times H(112) \times D(64))$ are generated from all four branches of STCF block. And the final STCF feature vector is obtained by stacking STCF branch wise convolution features as $STCF_{FV} = \{STCFV_{FVi}\}$, where $i\epsilon[1,4]$ and each branch vector is constructed with dimension $(W_i \times H_i \times D_i) = (112 \times 112 \times 64)$ that results in $STCF_{FV}$ with $[448 \times 448 \times 64]$ dimesion.

#### B. Attention Driven Residual Block (ADRB)

The key structure of residual units allows direct signal propagation from first to the last layer of the network and gradients to propagate from the loss layer to any previous layer by skipping the intermediate weight layers during backpropagation, which helps to handle the vanishing gradient problem to a great extent. Hence, the idea of residuals [35] proliferated the performance of deep networks by adding the identity function. The effect can be further enhanced by adding salient features instead of the identity function directly. It is implemented by using Attention Driven Residual Block (ADRB), shown in Fig. 3(a) and (b), where the attention block [36] tends to extract a spatial attention map by utilizing the inter-spatial relationship of features. The residual units are defined as follows:

$$y = \sigma_1(x + \mathcal{F}(x; \mathcal{W})) \quad (2)$$

where $x$ and $y$ are the input and output of the RIAC-Net architecture, $\sigma_1(.) \equiv$ ReLU [37], and $\mathcal{F}$ is a non-linear residual mapping function for input $x$ which is formulated as follows:

$$\mathcal{F}(x; \mathcal{W}) = STCF_{FV}(x; \mathcal{W}_k) \quad (3)$$

where is $\mathcal{W} = \{\mathcal{W}_k, 4 \leq k \leq 1\}$ for each convolution branch of STCF block. The residual unit is modified as

$$y' = \sigma_1(\Psi(x) + \mathcal{F}(x; \mathcal{W})) \quad (4)$$

And $\Psi(x) = x * \sigma_2\{f^{1\times1}\{\sigma_1[f^{7\times7}(x) + f^{1\times1}(\Lambda(x))]\}\}, \Psi \epsilon \mathbb{R}^{W \times H}$ (5)

where $\sigma_1$ is a ReLU function and $\sigma_2$ is a sigmoid function, $f^{1\times1}(.), f^{7\times7}(.)$ are non-linear convolution layers with $1 \times 1$ and $7 \times 7$ kernel sizes and $\Lambda(.)$ is a 2D max-pool layer with $2 \times 2$ kernel size. The '*' and '+' are multiplicative [38] and additive [39] attention operator. To extract salient features additive attention method performs better for large dimensional input features [40] whereas the multiplicative attention method holds fast computations and also memory-efficient due to the matrix multiplication. Therefore, at the input stage additive attention operator is applied to handle larger input dimension than the later one, as shown in Fig. 3 (c). The two different sized kernels in the convolution layer $f^{k \times k}$ cover the features on the coarse spatial grid level and finer grid level which collectively helps to identify relevant features and disambiguate the task irrelevant features in $x$.

Table 1: Description RIAC-Net –STCF Block Architecture parameters

| RIAC-Net Branches | No. of filters | Kernel size/stride | Input size $(W_I \times H_I \times D_I)$ | Output size $(W_O \times H_O \times D_O)$ |
|---|---|---|---|---|
| Branch 1 | Conv, 64 | $(1 \times 1)/2$ | $(224 \times 224 \times 3)$ | $(112 \times 112 \times 64)$ |
| Branch 2 | Conv, 32 | $(1 \times 1)/1$ | $(224 \times 224 \times 64)$ | $(224 \times 224 \times 32)$ |
| | Conv, 64 | $(3 \times 3)/2$ | $(224 \times 224 \times 32)$ | $(112 \times 112 \times 64)$ |
| Branch 3 | Conv, 128 | $(1 \times 1)/1$ | $(224 \times 224 \times 3)$ | $(224 \times 224 \times 128)$ |
| | Conv, 64 | $(3 \times 3)/1$ | $(224 \times 224 \times 128)$ | $(224 \times 224 \times 64)$ |
| | Conv, 64 | $(3 \times 3)/2$ | $(224 \times 224 \times 64)$ | $(112 \times 112 \times 64)$ |
| Branch 4 | Maxpool | $(2 \times 2)/1$ | $(224 \times 224 \times 3)$ | $(112 \times 112 \times 3)$ |
| | Conv, 64 | $(1 \times 1)/1$ | $(112 \times 112 \times 3)$ | $(112 \times 112 \times 64)$ |

Sample images of the residue and attention driven residue activation maps of the action, for complete CASS and part wise CASS, are shown in Fig. 4. It is clearly observed that the residue branch extracts noisy data whereas attention driven residue branch highlight the salient information about the action resulting in improved recognition performance.

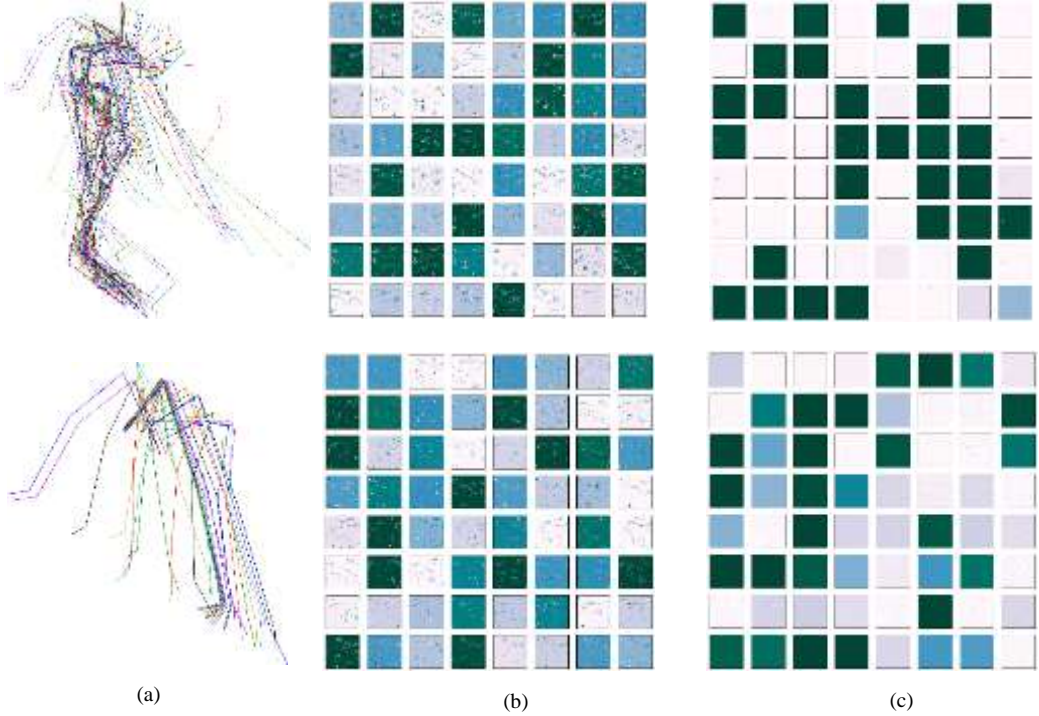

(a)   (b)   (c)

Fig. 4. Illustration of Residue and Attention driven residue activation maps (8 × 8) for complete CASS and left leg (LL) CASS (a) input CASS (b) residue activation maps (c) attention driven residue activation map skeleton human action recognition.

## C. Learning of part-wise RIAC-Net based action descriptors

Late fusion approach works better than the early fusion scheme [41] at the cost of additional learning attempts. Therefore, RIAC-Net based action descriptor is designed and learnt for each part-wise CASS, individually, using the combination of global average pooling (GAP), batch normalisation (BN), Long Short Term Memory (LSTM) layers, dropout layer with 0.2 dropout probability and dense layer. The final prediction is given by fusing the learnt part-wise CASS based predictions using weighted fusion scheme, as shown in Fig. 5. The best recognition performance for a specific weight combination i.e. $\{w_i\} \exists i \in [1,5]$, is reported finally. To learn the unique patterns of the part wise RIAC-Net based action features two consecutive LSTM layers are used in such a way that output gate $h_t^2$ of former LSTM layer is fed to the input gate $i_t^2$ of later LSTM layer. Let $x_t^i, h_t^i$, and $C_t^i$ be input, output and cell state of the $i^{th}$ LSTM layer, at instance $t$ respectively. The sequence of flow of the signal from the first LSTM layer to second layer is given by Eq. 6 to 9.

$$h_t^1 = \sigma(W_0^1 * [h_{t-1}^1, x_t^1] + b_o) * \tanh(C_t^1) \qquad (6)$$

$$C_t^1 = C_{t-1}^1 * f_t^1 + \widehat{C_t^1} * i_t^1 \qquad (7)$$

$$x_t^2 = h_t^1 \text{ and } h_t^2 = \sigma(W_0^2 * [h_{t-1}^2, x_t^2] + b_o) * \tanh(C_t^2) \qquad (8)$$

$$C_t^2 = C_{t-1}^2 * f_t^2 + \widehat{C_t^2} * i_t^2 \qquad (9)$$

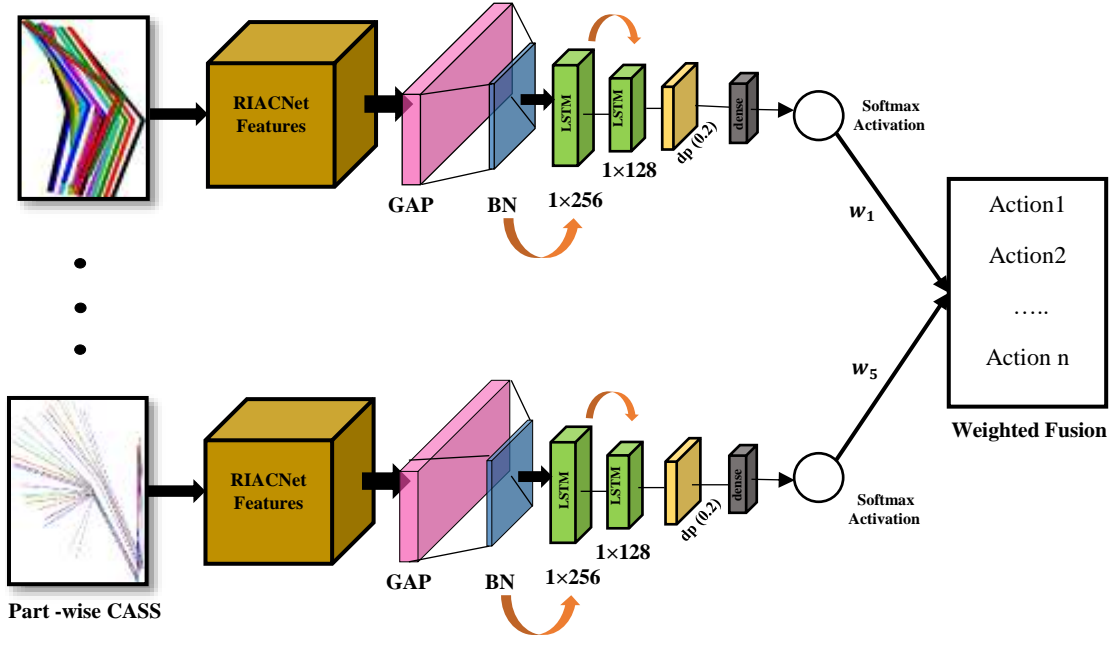

Fig. 5. Description of proposed Part-wise Spatiotemporal and Attention Driven Residues based Learning for skeleton human action recognition.

where $\sigma$ is the sigmoid operation, $W_0^i$ is the weights of the output gate of the $i^{th}$ LSTM layer, cell state $C_t^i$ of the $i^{th}$ layer is computed using both forget gate output as $f_t^i$ and input gate output as $i_t^t$. $\widehat{C_t^i}$ control the amount of update required to current cell state $C_t^i$, according to the input $x_t^i$ passed through input gate, $i_t^i$. The learnt vector $h_t^2$ with $[1 \times 128]$ dimension is fed to dropout layer to handle the overfitting problem followed by a dense layer and Softmax activation function. Predictions from each part-wise branch are fused using weighted fusion that utilises all possible combination of weights $w_i$ to find the best set of weights, as shown follows:

$$P_c = \sum_{i=1}^{5} w_i \times p_i, c \in (1, n) \qquad (10)$$

where n is the number of classes.

## 4. Experiments and Results

To validate the performance of the proposed framework for skeleton-based human action recognition, three publically available 3D datasets- UT Kinect Action 3D and Florence 3D actions Dataset and MSR Daily Action3D datasets.

*4.1 UT Kinect Action 3D dataset*

UT Kinect Action 3D dataset [42] captures 10 types of human actions in indoor settings, using single stationary Kinect hardware. The dataset includes an RGB image with 640×480 resolution, depth image with 320×240 and twenty 3D joints of a human skeleton per frame captured at 30 FPS. The 10 actions include i) walk, ii) sit down iii) stand up iv) pick up v) carry vi) throw vii) push viii) pull ix) wave and x) clap hands. These actions are performed by 10 different actors (9 males and 1 female), each repeated two times. Hence, it consists of a total 200 (10×10×2) action samples with 6220 frames. The challenge lies in the fact that there exist viewpoint variations and high intra-class variations. The length of the sample actions ranges from 5 frames to 120 frames. Therefore, in the proposed work the number of frames for each action sample is made equal to 60 frames, before generating CASS representation of the action. It is performed by down-sampling the frames of the actions which possess more than 60 frames. And up-sampling is applied to the actions which possess less than 60 frames to maintain a symmetricity in the CASS generated using action frames. The sample RGB images are shown in Fig. 6 below. We use the skeleton representation of human actions for human action recognition.

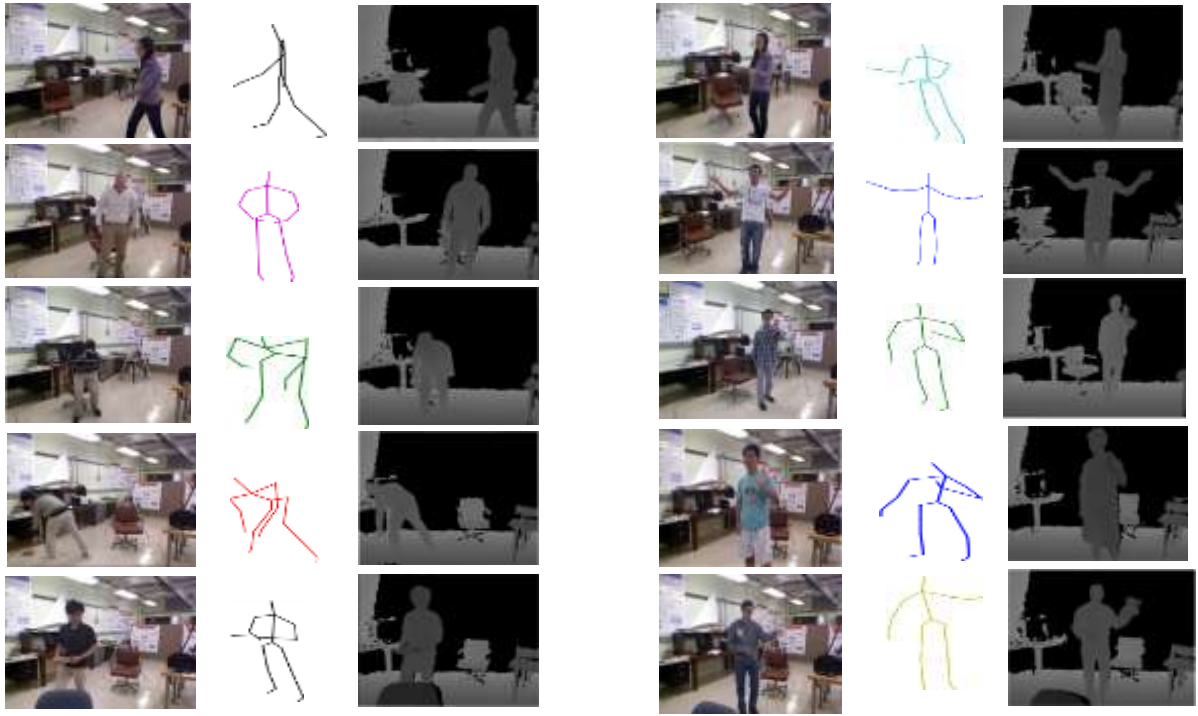

Fig. 6. Sample images of the publically available UT-Kinect Action 3D Dataset

*4.2 Florence 3D Action dataset*

Florence 3D Action dataset [43] dataset is collected at the University of Florence using a Kinect camera. It includes 9 actions: i) arm wave, ii) drink from a bottle, iii) answer phone, iv) clap, v) tight lace, vi) sit down, vii) stand up, viii) read watch, ix) bow, as shown in Fig. 7. Each action is performed by 10 subjects several times for a total of 215 sequences. The sequences are acquired using the OpenNI SDK, with skeletons represented by 15 joints instead of 20 as with the Microsoft Kinect SDK. The main challenges of this dataset are the similarity between actions, the human object interaction, and the different ways of performing the same action.

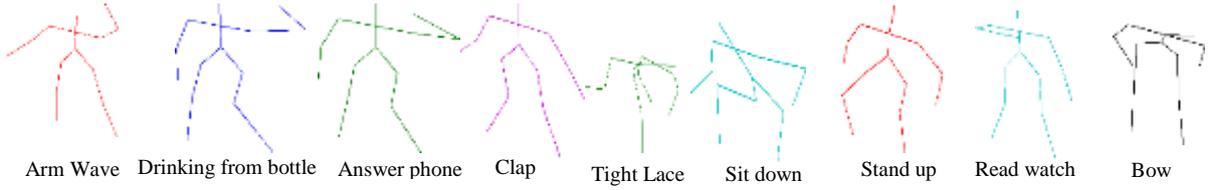

Arm Wave　Drinking from bottle　Answer phone　Clap　Tight Lace　Sit down　Stand up　Read watch　Bow

Fig. 7. Sample frames of Florence Action 3D dataset

### *4.3 MSR Action 3D Dataset*

The MSR Action 3D dataset [44] consists of 20 different action classes. Each action is performed by 10 subjects for three times. It is divided into 3 subsets AS1, AS2, AS3 each include 8 actions, as shown in Table 2. It includes 567 skeleton sequences. However, skeletons for 10 sequences are missing. Therefore, the experiments are conducted on 557 sequences, each provided with 3D coordinates of 20 joints per frame. For experimentation, cross-subject evaluation protocol is used. According to which, half of the dataset with 1, 3, 5, 7 and 9 subject ids, is used for training and another half of the dataset with 2, 4, 6, 8, 10 subject ids is used for testing for each action set-AS1, AS2, AS3.

Table 2: List of action classes in each subset AS1, AS2, and AS3 of the MSR Action 3D dataset

| Activity set 1 (AS1) | Horizontal arm wave | Hammer | Forward punch | High Throw | Hand clap | Bend | Tennis serve | Pick up and throw |
|---|---|---|---|---|---|---|---|---|
| Activity set 2 (AS2) | Horizontal arm wave | Hand catch | Draw X | Draw Tick | Draw Circle | Two hand wave | Forward kick | Side boxing |
| Activity set 3 (AS3) | High throw | Forward kick | Side kick | Jogging | Tennis Swing | Tennis Serve | Golf Swing | Pick up and Throw |

*Data augmentation:* Deep neural networks demand a large amount of training data to perform efficiently. We have only 557,428, and 200 skeleton sequences, for MSR Action 3D dataset, Florence 3D Action dataset and UT Kinect Action 3D Dataset. Thus, to prevent overfitting, data augmentation techniques are applied before processing the data. It includes cropping, horizontal and vertical flip, rotation at $45^0$ and $-45^0$.

The RIAC-Net architecture is implemented and evaluated in Python with Keras framework using the Tensor-Flow backend. We used mini-batches of 256 images, and Adam optimizer with default parameters, β1 = 0.9 and β2 = 0.999, during training. The initial learning rate is set to 0.001 and is decreased by a factor of 0.02 after every 20 epochs. The network is trained for each set of part wise input (HS, LH, RH, LL, RL), for 1000 epochs from scratch. To handle the overfitting in the training phase, we adopted weight noise and early stopping [45] along with drop-out strategy.

Table 3: Performance of the proposed framework for UT Kinect dataset

|  | Full skeleton | Head Spinal | Left Leg | Right Leg | Left hand | Right hand | Weighted Fusion |
|---|---|---|---|---|---|---|---|
| Training Loss | 0.2836 | 0.4480 | 0.4697 | 0.3209 | 0.3164 | 0.3801 | $W_{HS}, W_{LL}, W_{RL}, W_{LH}, W_{RH}$ |
| Training Accuracy | 99.96 | 92.94 | 92.94 | 97.65 | 97.13 | 95.92 | {2,3,4,4,5} |
| Test accuracy | 97.71 | 97.49 | 97.49 | 96.45 | 95.94 | 96.48 | **100.00** |

Table 4: Performance of the proposed framework for Florence 3D Action Dataset

| | Full skeleton | Head Spinal | Left Leg | Right Leg | Left hand | Right hand | Weighted Fusion |
|---|---|---|---|---|---|---|---|
| Training Loss | 0.1221 | 0.3700 | 0.0279 | 0.0630 | 0.0868 | 0.0127 | $W_{HS}, W_{LL}, W_{RL}, W_{LH}, W_{RH}$ |
| Training Accuracy | 100.00 | 100.00 | 100.00 | 99.69 | 96.65 | 100.00 | {3,4,2,3,2} |
| Test accuracy | 95.89 | 100.00 | 92.47 | 95.89 | 92.63 | 91.85 | **98.33** |

The experimental results on UT Kinect 3D Action dataset, Florence 3D Action dataset and MSR Action 3D dataset are reported in Table 3, 4 and 5 respectively. The weights $w_i, i\epsilon(1,5)$ corresponding with the best test accuracy achieved are also provided in Table 3, 4 and 5. The Validation loss curves for UT Kinect 3D Action dataset, Florence 3D Action dataset and MSR Action 3D dataset are shown in Fig. 7 (a), (b), and (c)-(e), respectively.

Table 5: Performance of the proposed framework for MSR action 3d dataset under cross-subject evaluation strategy

| Dataset subsets | Full skeleton | Head Spinal | Left Leg | Right Leg | Left hand | Right hand | Weighted Fusion Accuracy $W_{HS}, W_{LL}, W_{RL}, W_{LH}, W_{RH}$ |
|---|---|---|---|---|---|---|---|
| AS1 | 94.8 | 96 | 95 | 98.7 | 96 | 90 | 96.7 { **2,3,3,3,3** } |
| AS2 | 93.3 | 97.33 | 100.00 | 99.5 | 96.44 | 93.3 | **98.6** { **1,5,4,1,5** } |
| AS3 | 96.5 | 100 | 98.2 | 100 | 97.3 | 95.4 | **99.9** { **2,4,1,1,3** } |
| Overall | 94.8 | 97.77 | 96.84 | 99.39 | 96.58 | 92.9 | **98.40** |

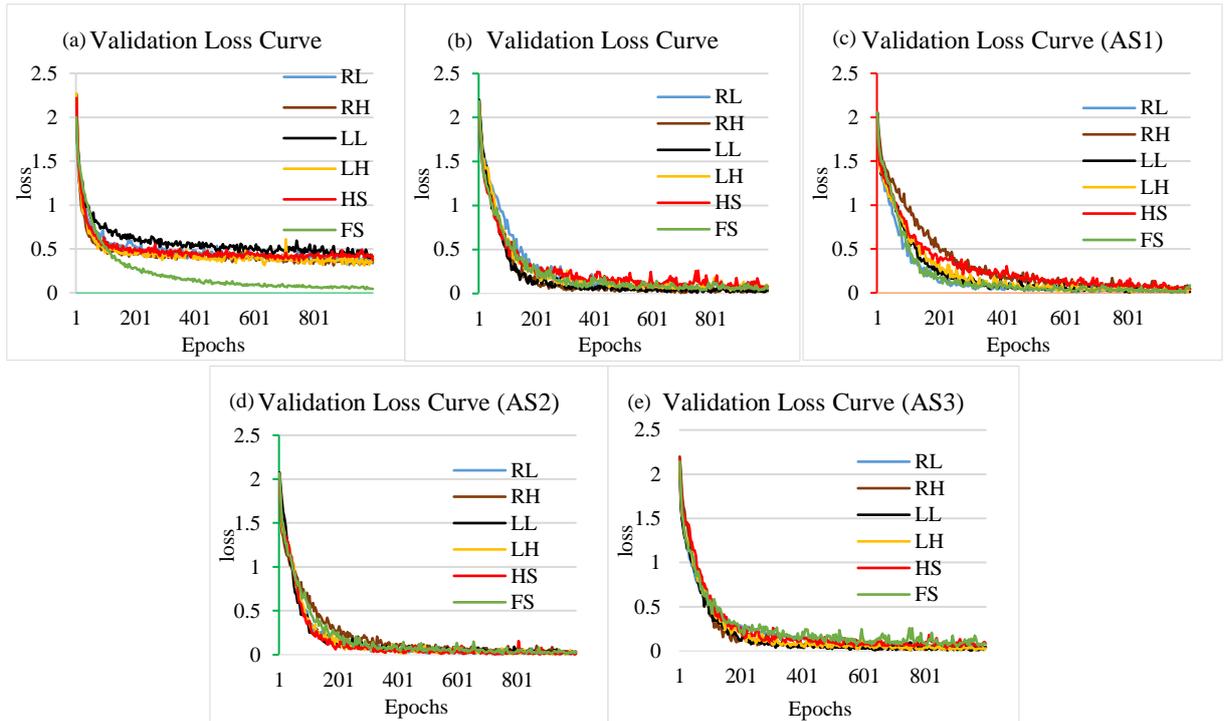

Fig.7. Illustration of Part-Wise (RL, RH, LL , LH, HS) and full skeleton (FS) based validation loss curves for (a) UT Kinect 3D Action Dataset, (b) Florence 3D Action dataset, (c)-(e) MSR Action 3D dataset AS1, AS2, and AS3

The Validation losses gradually decrease with the epochs, which confirms the adequate learning of the models for each part. Receiver output Characteristic (ROC) curve plotted between True Positive Rate (TPR) and False Positive Rate (FPR), in Fig.8 (a) and (b), also support the fact that weighted fusion of part-wise skeleton RIAFNet features turned out better action representation than Full skeletons (FS) for UT Kinect 3D Action dataset, Florence 3D Action dataset and MSR Action 3D dataset. Area Under the Curve (AUC), i.e. $AUC \in (0,1)$, is also computed for each method. The highest AUC values 1.00, 0.97, (0.97,0.99, and 1.00) are obtained for weighted average late fusion strategy over FS and part-wise (HS, LL, LH, RH, RL) skeleton based approaches for UT Kinect 3D Action dataset, Florence 3D Action dataset, MSR Action 3D dataset-AS1, AS2, AS3 sets respectively.

Table 6: A comparison of the proposed framework with other state-of-the-arts for UT Kinect Action dataset

| Method | Learning Model | Protocol | Accuracy (%) |
|---|---|---|---|
| Feature Combination [46] | K-NN | LOOCV | 98 |
| ST-LSTM+Trust Gate [20] | Hierarchal RNN | LOOCV | 97 |
| Grassmann Manifold [47] | LTBSVM | LOOCV | 88.5 |
| Geometric Features [11] | Multi-layer LSTM | cross validation | 95.96 |
| TS-LSTM [48] | Ensemble Temporal sliding LSTM | cross validation | 96.97 |
| Lie groups [7] | SVM | One vs all cross validation | 97.08 |
| Kernel Linearization [49] | SVM | cross validation | 98.2 |
| LRCNLG [30] | LSTM | LOCCV | 98.5 |
| **Proposed work** | **LSTM layers** | **LOOCV** | **100.00** |

Table 7: A comparison of the proposed framework with other state-of-the-arts for Florence 3D Action dataset

| Method | Learning Model | Protocol | Accuracy (%) |
|---|---|---|---|
| Lie groups [7] | SVM | One vs all Cross validation | 90.88 |
| Kernel Linearization [49] | SVM | Cross validation | 95.23 |
| Riemannian Manifold [50] | K-NN | LOOCV | 87.04 |
| Mining key pose [51] | Inference Algorithm | LOOCV | 92.25 |
| Feature combination [46] | K-NN | LOOCV | 94.39 |
| LRCNLG [30] | LSTM | LOOCV | 95.37 |
| **Proposed work** | **LSTM layers** | **LOOCV** | **98.33** |

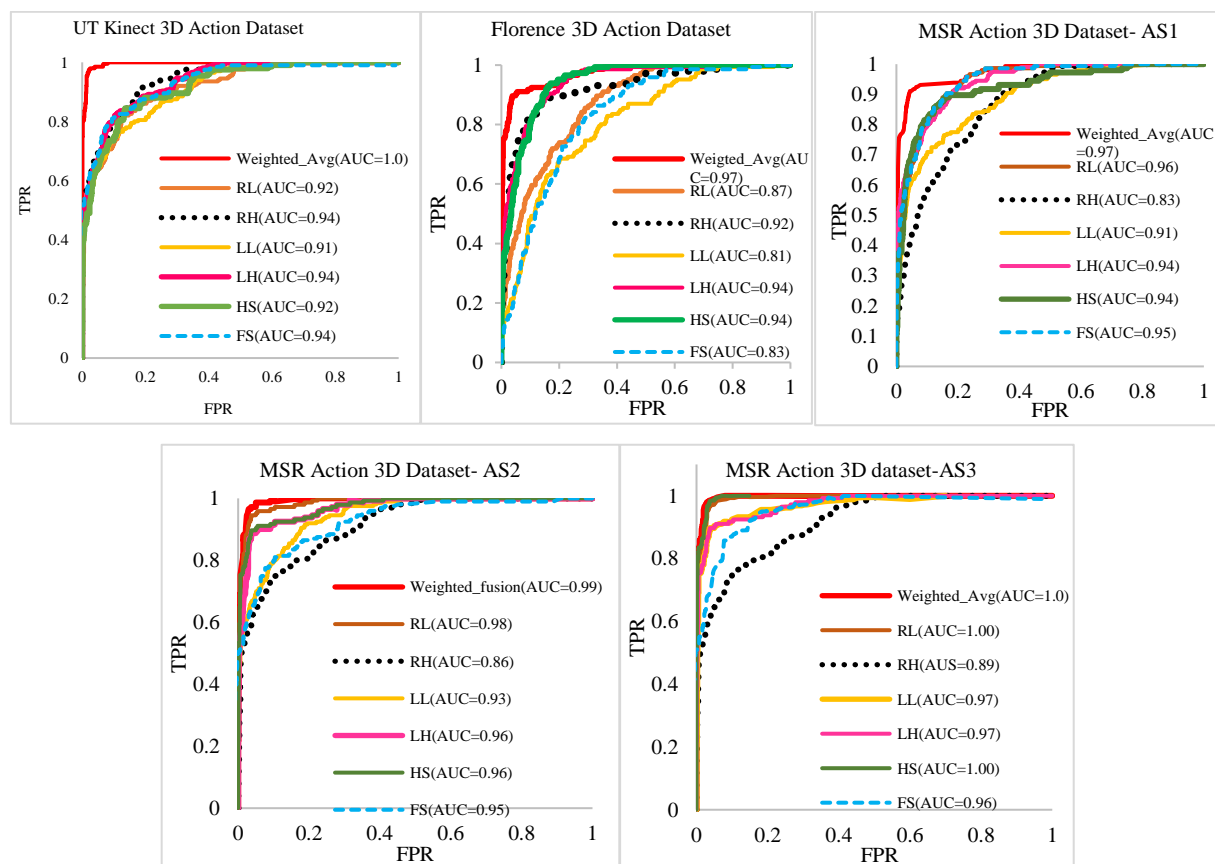

Fig. 8. ROC Curves of a) UT Kinect 3D dataset b) Florence Action 3D dataset c) MSR Action 3D dataset.

The achieved accuracy of the proposed work is compared with the other-state-of-arts for UT Kinect 3D Action dataset, Florence 3D Action dataset and MSR Action 3D dataset in Table 6, 7, and 8. The obtained results outperforms many previous studies [7]- [30], [10] - [14]. The proposed work achieved 100% recognition accuracy for UT Kinect 3D Action dataset using Leave-One-Out Cross Validation (LOOCV) scheme.

**Table 8: A comparison of the proposed framework with other state-of-the-arts for MSR Action 3D dataset cross subject evaluation**

| Method | Learning Model | AS1 | AS2 | AS3 | Average Accuracy (%) |
|---|---|---|---|---|---|
| HBRNN-L [10] | Hierarchical RNN | 93.33 | 94.64 | 95.50 | 94.49 |
| ST-NBNN [33] | Naive-Bayes Nearest-Neighbor | 91.50 | 95.60 | 97.30 | 94.80 |
| DMM-LBP-DF [52] | K-ELM | 98.10 | 92.00 | 94.60 | 94.90 |
| VBDDM [53] | ProCRC | 99.10 | 92.30 | 98.20 | 96.50 |
| Mean3DJ [16] | Random Forest | - | - | - | 82.68 |
| Lie group-MinP-PrefixSpan [54] | SVM | - | - | - | 97.4 |
| SPMFs [15] | D-CNN | 97.06 | 99.00 | 98.09 | 98.05 |
| ResNet-44 [14] | D-CNN | 99.90 | 99.80 | 100 | 99.90 |
| **Proposed method** | **LSTM layers** | **95.70** | **98.60** | **99.90** | **98.06** |

The Classification result for each AS1, AS2, and AS3 sets of the MSR-Action3D dataset are shown in Fig. 9 (c), (d), and (e). It is noticed that misclassification occurs only for the actions with high inter class similarity such as 'draw tick' and 'draw *X*', 'Pickup and Throw' and 'Bend'. Whereas "Forward Kick" and "Tennis Serve" actions which share a large overlap in the sequences, are more challenging to distinguish the two actions in AS3 set. The proposed framework handled this inter class similarity between the two actions and recognised 'Forward Kick' and 'Tennis Serve' with 100% accuracy.

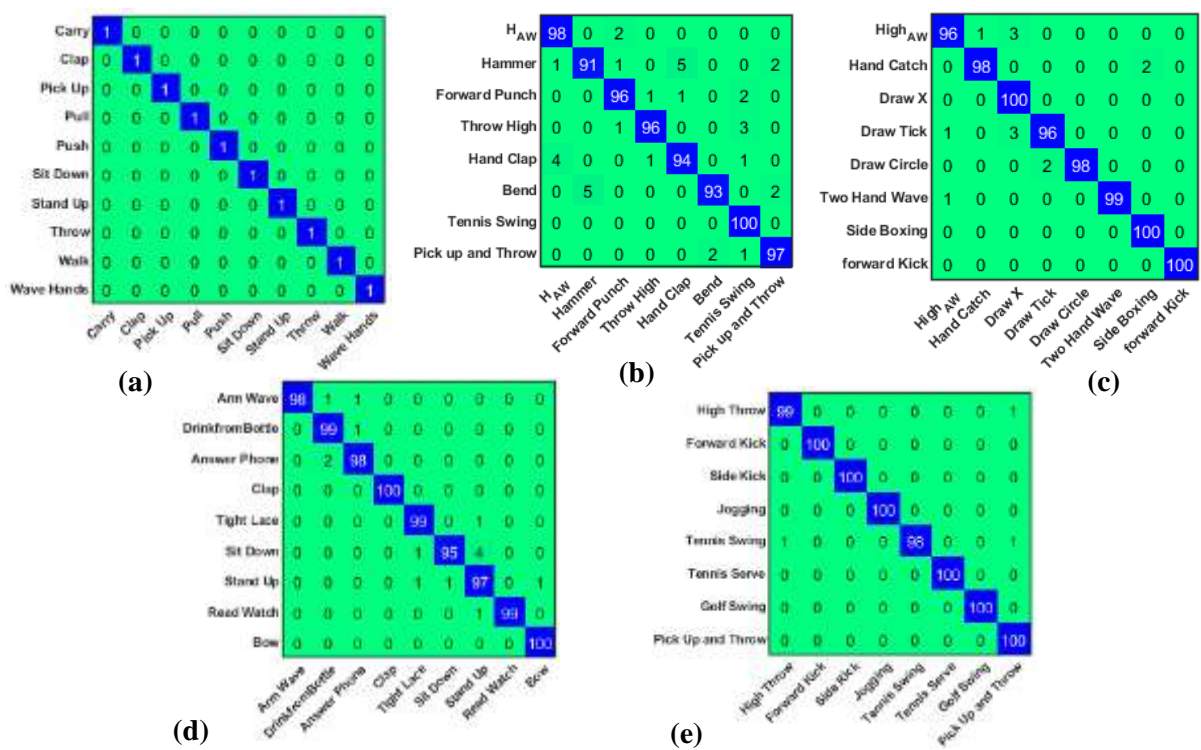

Fig. 9. Confusion Matrix of the a) UT Kinect Dataset, b) Florence Action 3D dataset and MSR action 3D dataset c) AS1, d) AS2 and e) AS3 sets. Where H_A_W: Horizontal Arm Wave and High_A_W: High Arm Wave

The weighted classification confusion matrix of the UT Kinect 3D action dataset is shown in Fig. 9 (a). From where it is clearly evident that each action is recognized correctly without any misclassification irrespective of the presence of high intra class variations and view variations.

The obtained result outperforms many previous studies Lie groups [7] , LRCNLG [30] , Grassmann Manifold [47], TS-LSTM [48] which tried to learn geometrical 3D features of human actions using Lie groups, Grassmann Manifold and temporal sliding LSTMs respectively.

The proposed work achieved 98.33% recognition accuracy on Florence 3D Action dataset, which is 2.96% higher than LRCNLG [30] that integrated Lie groups with deep neural networks to learn the geometrical 3D features. The confusion matrix for Florence 3D action dataset is shown in Fig. 9 (b) which shows that the proposed work obtained very high accuracy for most of the actions. However, there exist some confusion between similar actions such as 'Answer Phone', 'drink from bottle', and 'high arm wave', 'stand up' and 'sit down'. We have achieved 98.05% fairly a high recognition accuracy on MSR action 3D dataset which outperformed previous works [15] [33]. However, Pham et al. [14] achieved 99.90% accuracy which utilised deep ResNets to process skeleton data for human action recognition. Some skeleton based methods like [33] [15] used skeleton features based on pairwise distances between joints. However, our results obtained on MSR action 3-D dataset show that part wise analysis of whole skeletons followed by late fusion approach is more discriminative approach than taking into consideration the joints separately.

## 5. Conclusion

In this paper, an effective skeleton based part-wise spatio-temporal CNN - RIAFNetwork based 3D human action recognition framework is proposed. It models the dynamics of the action by splitting the skeletons into five parts- Head to Spine (HS), Left Leg (LL), Right Leg (RL), Left Hand (LH), Right Hand (RH). Each part of the skeleton behaves differently for every action which is encrypted using RIAC-Net network which helps to highlight local dynamics {LL, LH, RH, RL, HS} of the action, that proved superior representation than the global action dynamics {FS} of the skeleton. The architecture of the RAIC-Net is designed using the concept of attention based residues and inception blocks. The final action class scores are generated by weighted (decision level) fusion of deep features. The empirical results and the analysis of the performance of our proposed approach exhibit promising results with high accuracies 100%,98.03%, and 98.7% on UT Kinect Action 3D and Florence 3D actions Dataset and MSR Daily Action3D datasets. Obtained results show that weighted fusion of part wise skeleton action dynamics' learning performs better than FS based action recognition. It is also observed that the proposed model is able to handle the intra class variations and inter class similarity among the actions quite decently.

In future, the work can be extended to be capable enough to handle view variations and partial occlusions in real-time scenario, which are the two very challenging aspects for a robust action recognition system.